\documentclass[11pt,a4paper]{article}
\usepackage[hyperref]{acl2021}
\usepackage{times}
\usepackage{latexsym}

\usepackage{microtype}

\aclfinalcopy 

\def\aclpaperid{1393}

\usepackage[T1]{fontenc}

\usepackage[utf8]{inputenc}

\usepackage{microtype}
\usepackage{graphicx}
\usepackage{amsmath}

\usepackage[export]{adjustbox}
\usepackage{array}

\usepackage{enumitem}

\usepackage[ruled,vlined]{algorithm2e}
\DeclareMathAlphabet{\pazocal}{OMS}{zplm}{m}{n}

\usepackage{graphicx}               
\usepackage{tabularx}               
\newcolumntype{C}{>{\centering\arraybackslash}X}
\usepackage{multirow}               
\usepackage{diagbox}                
\usepackage{hhline}                 
\usepackage{color}                  
\usepackage{amssymb}                
\usepackage{mathtools}              
\usepackage{enumitem}               

\usepackage{makecell}
\usepackage{xcolor}

\definecolor{effectspancolor}{RGB}{0, 51, 125}
\definecolor{drugspancolor}{RGB}{44, 7, 110}

\usepackage{xparse}
\NewDocumentCommand{\heng}{ mO{} }{\textcolor{red}{\textsuperscript{\textit{Heng}}\textsf{\textbf{\small[#1]}}}}
\NewDocumentCommand{\tuan}{ mO{} }{\textcolor{blue}{\textsuperscript{\textit{Tuan}}\textsf{\textbf{\small[#1]}}}}
\NewDocumentCommand{\zixuan}{ mO{} }{\textcolor{orange}{\textsuperscript{\textit{Zixuan}}\textsf{\textbf{\small[#1]}}}}

\newcommand{\cev}[1]{\reflectbox{\ensuremath{\vec{\reflectbox{\ensuremath{#1}}}}}}

\title{Joint Biomedical Entity and Relation Extraction with Knowledge-Enhanced Collective Inference}

\author{Tuan Lai \textsuperscript{1}, Heng Ji \textsuperscript{1}, \textbf{ChengXiang Zhai \textsuperscript{1}}, Quan Hung Tran \textsuperscript{2}\\
	    \textsuperscript{1}University of Illinois at Urbana-Champaign
	    \textsuperscript{2}Adobe Research\\
        \{tuanml2, hengji, czhai\}@illinois.edu\\  qtran@adobe.com
}

\date{}

\begin{document}
\maketitle

\begin{abstract}

Compared to the general news domain, information extraction (IE) from biomedical text requires much broader domain knowledge. However, many previous IE methods do not utilize any external knowledge during inference. Due to the exponential growth of biomedical publications, models that do not go beyond their fixed set of parameters will likely fall behind. 
Inspired by how humans 
look up relevant information to comprehend a scientific text, we present a novel framework that utilizes external knowledge for joint entity and relation extraction named \textbf{KECI} (Knowledge-Enhanced Collective Inference). Given an input text, KECI first constructs an initial span graph representing its initial understanding of the text. It then uses an entity linker to form a knowledge graph containing relevant background knowledge for the the entity mentions in the text. To make the final predictions, KECI fuses the initial span graph and the knowledge graph into a more refined graph using an attention mechanism. KECI takes a collective approach to link mention spans to entities by integrating global relational information into local representations using graph convolutional networks. Our experimental results show that the framework is highly effective, achieving new state-of-the-art results in two different benchmark datasets: BioRelEx (binding interaction detection) and ADE (adverse drug event extraction). For example, KECI achieves absolute improvements of 4.59\% and 4.91\% in F1 scores over the state-of-the-art on the BioRelEx entity and relation extraction tasks \footnote{The code is publicly available at \url{https://github.com/laituan245/bio_relex}}.
\end{abstract}

\section{Introduction}


With the accelerating growth of biomedical publications, it has become increasingly challenging to manually keep up with all the latest articles. As a result, developing methods for automatic extraction of biomedical entities and their relations has attracted much research attention recently \cite{li2017neural,10.1093/bioinformatics/btaa993,Luo2020ANN}. Many related tasks and datasets have been introduced, ranging from binding interaction detection (BioRelEx) \cite{khachatrian2019biorelex} to adverse drug event extraction (ADE) \cite{adedataset}.

\begin{figure}[!t]
\centering
\includegraphics[width=\linewidth]{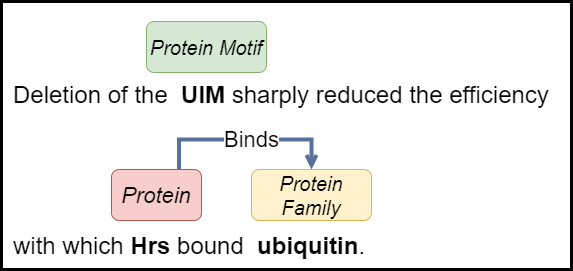}
\caption{An example in the BioRelEx dataset. \textbf{UIM} is an abbreviation of ``Ubiquitin-Interacting Motif''. Our baseline SciBERT model incorrectly predicts the mention as a ``DNA'' instead of a ``Protein Motif''.}
\label{fig:biorelex_intro_example}
\end{figure}

\begin{figure*}[!ht]
  \centering
  \includegraphics[width=\textwidth]{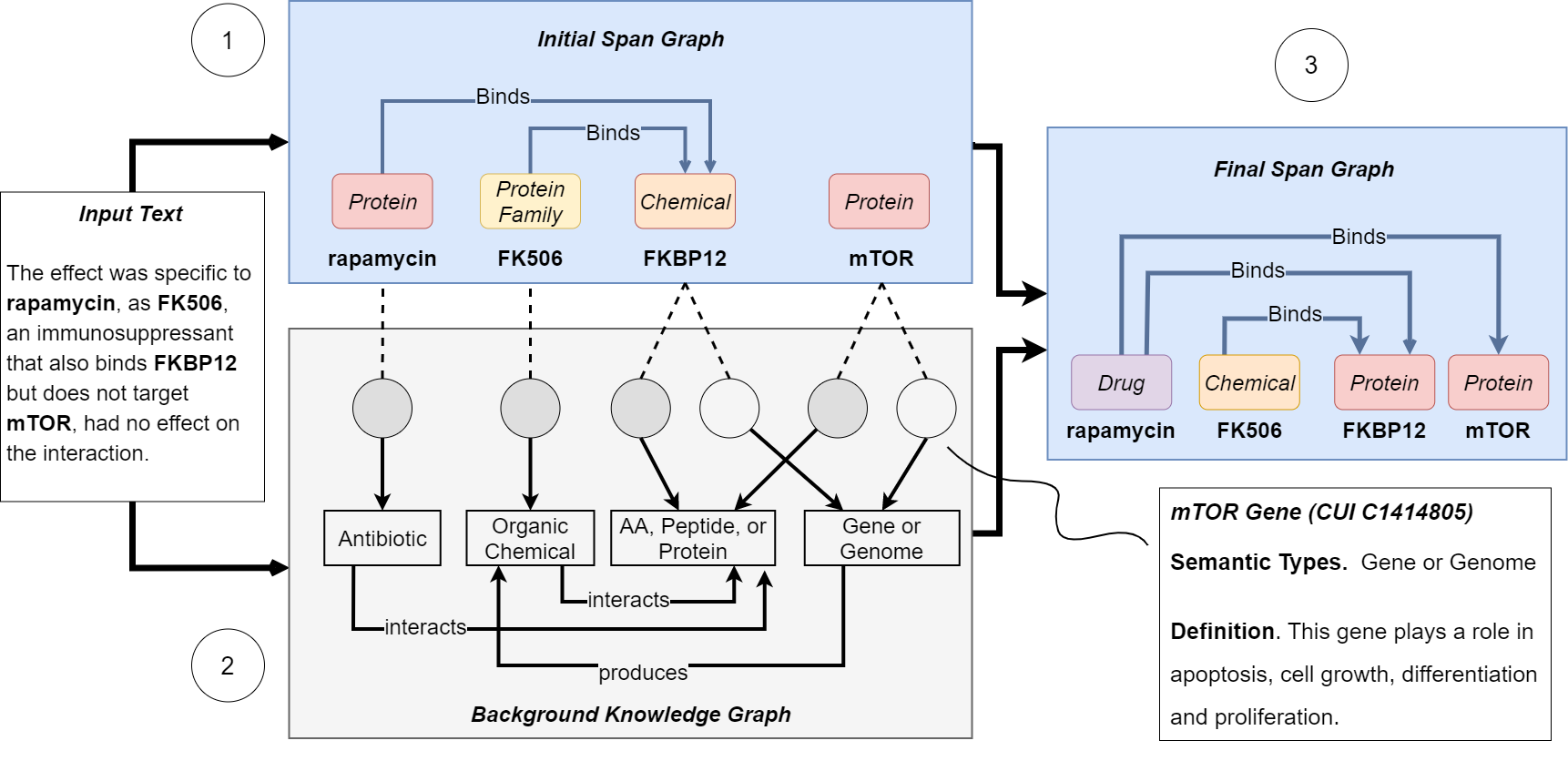}
  \caption{KECI operates in three main steps: (1) initial span graph construction (2) background knowledge graph construction (3) fusion of these two graphs into a final span graph. KECI takes a \textit{collective} approach to link multiple mentions simultaneously to entities by incorporating 
  \textit{global} relational information using GCNs.}
  \label{fig:approach_overview}
\end{figure*}

Many recent joint models for entity and relation extraction rely mainly on distributional representations and do not utilize any external knowledge source \cite{Eberts2020SpanbasedJE,jietal2020span,ijcai2020558}. However, different from the general news domain, information extraction for the biomedical domain typically requires much broader domain-specific knowledge. Biomedical documents, either formal (e.g., scientific papers) or informal ones (e.g., clinical notes), are written for domain experts. As such, they contain many highly specialized terms, acronyms, and abbreviations. In the BioRelEx dataset, we find that about 65\% of the annotated entity mentions are abbreviations of biological entities, and an example is shown in Figure \ref{fig:biorelex_intro_example}. These unique characteristics bring great challenges to general-domain systems and even to existing scientific language models that do not use any external knowledge base during inference \cite{Beltagy2019SciBERT,10.1093/bioinformatics/btz682}. For example, even though SciBERT \cite{Beltagy2019SciBERT} was pretrained on 1.14M scientific papers, our baseline SciBERT model still incorrectly predicts the type of the term \textit{UIM} in Figure \ref{fig:biorelex_intro_example} to be ``DNA'', which should be a ``Protein Motif'' instead. Since the biomedical literature is expanding at an exponential rate, models that do not go beyond their fixed set of parameters will likely fall behind.

In this paper, we introduce \textbf{KECI} (Knowledge-Enhanced Collective Inference), a novel end-to-end framework that utilizes external domain knowledge for joint entity and relation extraction. Inspired by how humans comprehend a complex piece of scientific text, the framework operates in three main steps (Figure \ref{fig:approach_overview}). KECI first reads the input text and constructs an initial \textit{span graph} representing its initial understanding of the text. In a span graph, each node represents a (predicted) entity mention, and each edge represents a (predicted) relation between two entity mentions. KECI then uses an entity linker to form a background knowledge graph containing all potentially relevant biomedical entities from an external knowledge base (KB). For each entity, we extract its semantic types, its definition sentence, and its relational information from the external KB. Finally, KECI uses an attention mechanism to fuse the initial span graph and the background knowledge graph into a more refined graph representing the final output. Different from previous methods that link mentions to entities based solely on \textit{local} contexts \cite{Li2020BiosemanticRE}, our framework takes a more collective approach to link multiple semantically related mentions simultaneously by leveraging global topical coherence. Our hypothesis is that if multiple mentions co-occur in the same discourse and they are probably semantically related, their reference entities should also be connected in the external KB. KECI integrates \textit{global} relational information into mention and entity representations using graph convolutional networks (GCNs) before linking.

The benefit of collective inference can be illustrated by the example shown in Figure \ref{fig:approach_overview}. The entity linker proposes two candidate entities for the mention \textit{FKBP12}; one is of semantic type ``AA, Peptide, or Protein'' and the other is of semantic type ``Gene or Genome''. It can be tricky to select the correct candidate as \textit{FKBP12} is already tagged with the wrong type in the initial span graph (i.e., it is predicted to be a ``Chemical'' instead of a ``Protein''). However, because of the structural resemblance between the mention-pair $\langle$\textit{FK506}, \textit{FKBP12}$\rangle$ and the pair $\langle$``Organic Chemical'', ``AA, Peptide, or Protein''$\rangle$, KECI will link \textit{FKBP12} to the entity of semantic type ``AA, Peptide, or Protein''. As a result, the final predicted type of \textit{FKBP12} will also be corrected to ``Protein'' in the final span graph.

Our extensive experimental results show that the proposed framework is highly effective, achieving new state-of-the-art biomedical entity and relation extraction performance on two benchmark datasets: BioRelEx \cite{khachatrian2019biorelex} and ADE \cite{adedataset}. For example, KECI achieves absolute improvements of 4.59\% and 4.91\% in F1 scores over the state-of-the-art on the BioRelEx entity and relation extraction tasks. Our analysis also shows that KECI can automatically learn to select relevant candidate entities without any explicit entity linking supervision during training. Furthermore, because KECI considers text spans as the basic units for prediction, it can extract nested entity mentions. 

\section{Methods}
\subsection{Overview}

KECI considers text spans as the basic units for feature extraction and prediction. This design choice allows us to handle nested entity mentions \cite{sohrabmiwa2018deep}. Also, joint entity and relation extraction can be naturally formulated as the task of extracting a \textit{span graph} from an input document \cite{luanetal2019general}. In a span graph, each node represents a (predicted) entity mention, and each edge represents a (predicted) relation between two entity mentions.

Given an input document $D$, KECI first enumerates all the spans (up to a certain length) and embeds them into feature vectors (Sec. \ref{sec:span_encoder}). With these feature vectors, KECI predicts an initial span graph and applies a GCN to integrate initial relational information into each span representation (Sec. \ref{sec:initial_span_graph}). KECI then uses an entity linker to build a background knowledge graph and applies another GCN to encode each node of the graph (Sec. \ref{sec:external_knowledge}). Finally, KECI aligns the nodes of the initial span graph and the background knowledge graph to make the final predictions (Sec. \ref{sec:final_graph}). We train KECI in an end-to-end manner without using any additional entity linking supervision (Sec. \ref{sec:training}).

Overall, the design of KECI is partly inspired by previous research in educational psychology. Students' background knowledge plays a vital role in guiding their understanding and comprehension of scientific texts \cite{Alvermann1985PriorKA,Braasch2010TheRO}.  ``Activating'' relevant and accurate prior knowledge will aid students’ reading comprehension.


\subsection{Span Encoder} \label{sec:span_encoder}
Our model first constructs a contextualized representation for each input token using SciBERT \cite{Beltagy2019SciBERT}. Let $\textbf{X} = (\textbf{x}_1, ..., \textbf{x}_n)$ be the output of the token-level encoder, where $n$ denotes the number of tokens in $D$. Then, for each span $s_i$ whose length is not more than $L$, we compute its span representation $\textbf{s}_i \in \mathbb{R}^{d}$ as:
\begin{equation} \label{eq:base_span_emb}
    \textbf{s}_i = \text{FFNN}_\text{g}\big(\big\lbrack\textbf{x}_{\text{START}(i)}, \textbf{x}_{\text{END}(i)}, \hat{\textbf{x}}_{i}, \phi(\textbf{s}_{i}) \big\rbrack\big)
\end{equation}
where $\text{START}(i)$ and $\text{END}(i)$ denote the start and end indices of $s_i$ respectively. $\textbf{x}_{\text{START}(i)}$ and $\textbf{x}_{\text{END}(i)}$ are the boundary token representations. $\hat{\textbf{x}}_{i}$ is an attention-weighted sum of the token representations in the span \cite{leeetal2017end}. $\phi(\textbf{s}_{i})$ is a feature vector denoting the span length. $\text{FFNN}_\text{g}$ is a feedforward network with $\mathtt{ReLU}$ activations.
\subsection{Initial Span Graph Construction} \label{sec:initial_span_graph}

With the extracted span representations, we predict the type of each span and also the relation between each span pair jointly. Let $E$ denote the set of entity types (including non-entity), and $R$ denote the set of relation types (including non-relation). We first classify each span $s_i$:
\begin{equation}\label{equa:baseline_entity}
\textbf{e}_{i} = \mathtt{Softmax}\big(\text{FFNN}_\text{$e$}(\textbf{s}_{i})\big)
\end{equation}
where $\text{FFNN}_\text{$e$}$ is a feedforward network mapping from $\mathbb{R}^{d} \rightarrow \mathbb{R}^{|E|}$. We then employ another network to classify the relation of each span pair $\langle s_i, s_j \rangle$:
\begin{equation}\label{equa:baseline_relation}
\textbf{r}_{ij} = \mathtt{Softmax}\big(\text{FFNN}_\text{$r$}\big(\big\lbrack \textbf{s}_i,\textbf{s}_j, \textbf{s}_i \circ \textbf{s}_j \big\rbrack\big)\big)
\end{equation}
where $\circ$ denotes the element-wise multiplication, $\text{FFNN}_\text{$r$}$ is a mapping from $\mathbb{R}^{3 \times d} \rightarrow \mathbb{R}^{|R|}$. We will use the notation $\textbf{r}_{ij}[k]$ to refer to the predicted probability of $s_i$ and $s_j$ having the relation $k$.

At this point, one can already obtain a valid output for the task from the predicted entity and relation scores. However, these predictions are based solely on the local document context, which can be difficult to understand without any external domain knowledge. Therefore, our framework uses these predictions only to construct an initial span graph that will be refined later based on information extracted from an external knowledge source.

To maintain computational efficiency, we first prune out spans of text that are unlikely to be entity mentions. We only keep up to $\lambda n$ spans with the lowest probability scores of being a non-entity. The value of $\lambda$ is selected empirically and set to be 0.5. Spans that pass the filter are represented as nodes in the initial span graph. For every span pair $\langle s_i, s_j \rangle$, we create $|R|$ directed edges from the node representing $s_i$ to the node representing $s_j$. Each edge represents one relation type and is weighted by the corresponding probability score in $\textbf{r}_{ij}$.

Let $G_s = \{V_s, E_s\}$ denote the initial span graph. We use a bidirectional GCN \cite{marcheggianititov2017encoding,fuetal2019graphrel} to recursively update each span representation:
\begin{equation}
\begin{split}
\vec{\textbf{h}_{i}^{l}} &= \sum_{s_j \in V_s \setminus \{s_i\}} \sum_{k \in R} \textbf{r}_{ij}[k] \bigg(\vec{\textbf{W}_k^{(l)}}\textbf{h}_{j}^{l} + \vec{\textbf{b}_k^{(l)}} \bigg) \\
\cev{\textbf{h}_{i}^{l}} &= \sum_{s_j \in V_s \setminus \{s_i\}} \sum_{k \in R} \textbf{r}_{ji}[k] \bigg(\cev{\textbf{W}_k^{(l)}}\textbf{h}_{j}^{l} + \cev{\textbf{b}_k^{(l)}}\bigg) \\
\textbf{h}_{i}^{l+1} &= \textbf{h}_{i}^{l} + \text{FFNN}_a^{(l)} \Bigg(\mathtt{ReLU}\bigg(\big\lbrack \vec{\textbf{h}_{i}^{l}}, \cev{\textbf{h}_{i}^{l}} \big\rbrack\bigg)\Bigg)
\end{split}
\end{equation}
where $\textbf{h}_{i}^{l}$ is the hidden feature vector of span $s_i$ at layer $l$. We initialize $\textbf{h}_{i}^{0}$ to be $\textbf{s}_i$ (Eq. \ref{eq:base_span_emb}). $\text{FFNN}_a^{(l)}$ is a feedforward network whose output dimension is the same as the dimension of $\textbf{h}_{i}^{l}$.

After multiple iterations of message passing, each span representation will contain the global relational information of $G_s$. Let $\textbf{h}_{i}$ denote the feature vector at the final layer of the GCN. Note that the dimension of $\textbf{h}_i$ is the same as the dimension of $\textbf{s}_i$ (i.e., $\textbf{h}_{i} \in \mathbb{R}^{d}$).

\subsection{Background Knowledge Graph Construction} \label{sec:external_knowledge}

In this work, we utilize external knowledge from the Unified Medical Language System (UMLS) \cite{Bodenreider2004TheUM}. UMLS consists of three main components: Metathesaurus, Semantic Network, and Specialist Lexicon and Lexical Tools. The Metathesaurus provides information about millions of fine-grained biomedical concepts and relations between them. To be consistent with the existing literature on knowledge graphs, we will refer to UMLS concepts as entities. Each entity is annotated with one or more higher-level semantic types, such as \textit{Anatomical Structure}, \textit{Cell}, or \textit{Virus}. In addition to relations between entities, there are also semantic relations between semantic types. For example, there is an \textit{affects} relation from \textit{Acquired Abnormality} to \textit{Physiologic Function}. This information is provided by the Semantic Network.

We first extract UMLS biomedical entities from the input document $D$ using MetaMap, an entity mapping tool for UMLS \cite{Aronson2010AnOO}. We then construct a background knowledge graph (KG) from the extracted information. More specifically, we first create a node for every extracted biomedical entity. The semantic types of each entity node are also modeled as type  nodes that are linked with associated entity nodes. Finally, we create an edge for every relevant relation found in the Metathesaurus and the Semantic Network. An example KG is in the grey shaded region of Figure \ref{fig:approach_overview}.  Circles represent entity nodes, and rectangles represent nodes that correspond to semantic types. 

Note that we simply run MetaMap with the default options and do not tune it. In our experiments, we found that MetaMap typically returns many candidate entities unrelated to the input text. However, as to be discussed in Section \ref{sec:attn_pattern}, we show that KECI can learn to ignore the irrelevant entities.

Let $G_k = \{V_k, E_k\}$ denote the constructed background KG, where $V_k$ and $E_k$ are the node and edge sets, respectively. We use a set of UMLS embeddings pretrained by \newcite{Maldonado2019AdversarialLO} to initialize the representation of each node in $V_k$. We also use SciBERT to encode the UMLS definition sentence of each node into a vector and concatenate it to the initial representation. After that, since $G_k$ is a heterogeneous relational graph, we use a relational GCN \cite{Schlichtkrull2018ModelingRD} to update the representation of each node $v_i$:
\begin{equation}
    \textbf{v}_{i}^{l+1} = \mathtt{ReLU}\Bigg(\textbf{U}^{(l)} \textbf{v}_{i}^{l} + \sum_{k \in R} \sum_{v_j \in N^{k}_{i}} \bigg(\frac{1}{c_{i,k}}  \textbf{U}^{(l)}_{k} \textbf{v}_{j}^{l} \bigg) \Bigg)
\end{equation}
where $\textbf{v}_i^l$ is the feature vector of $v_i$ at layer $l$. $N^{k}_{i}$ is the set of neighbors of $v_i$ under relation $k \in R$. $c_{i,k}$ is a normalization constant and set to be $|N^{k}_{i}|$.

After multiple iterations of message passing are performed, the global relational information of the KG will be integrated into each node's representation. Let $\textbf{v}_{i}$ denote the feature vector at the final layer of the relational GCN. We further project each vector $\textbf{v}_i$ to another vector $\textbf{n}_i$ using a simple feedforward network, so that $\textbf{n}_i$ has the same dimension as the span representations (i.e., $\textbf{n}_i \in \mathbb{R}^{d}$). 


\subsection{Final Span Graph Prediction} \label{sec:final_graph}

At this point, we have two graphs: the initial span graph $G_s = \{V_s, E_s\}$ (Sec. \ref{sec:initial_span_graph}) and the background knowledge graph $G_k = \{V_k, E_k\}$ (Sec. \ref{sec:external_knowledge}). We have also obtained a structure-aware representation for each node in each graph (i.e., $\textbf{h}_i$ for each span $s_i \in V_s$ and $\textbf{n}_j$ for each entity $v_j \in V_k$).

\begin{figure}[!t]
\centering
\includegraphics[width=0.7\linewidth]{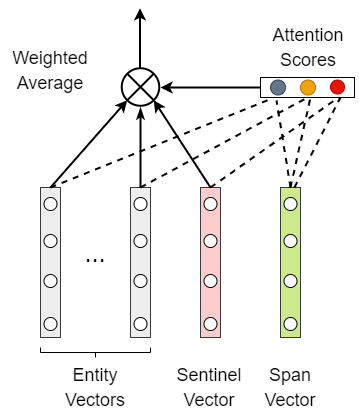}
\caption{An illustration of the attention mechanism.}
\label{fig:attention_mechanism}
\end{figure}

The next step is to soft-align the mentions and the candidate entities using an attention mechanism (Figure \ref{fig:attention_mechanism}). Let $C(s_i)$ denote the set of candidate entities for a span $s_i \in V_s$. For example, in Figure \ref{fig:approach_overview}, the mention \textit{FKBP12} has two candidate entities, while \textit{FK506} has only one candidate. For each candidate entity $v_j \in C(s_i)$, we calculate a scalar score $\alpha_{ij}$ indicating how relevant $v_j$ is to $s_i$:
\begin{equation}
  \alpha_{ij} = \text{FFNN}_\text{c}\big(\big[\textbf{h}_i, \textbf{n}_j\big]\big)
\end{equation}
where $\text{FFNN}_\text{c}$ is a feedforward network mapping from $\mathbb{R}^{2 \times d} \rightarrow \mathbb{R}$. Then we compute an additional sentinel vector $\textbf{c}_i$  \cite{yangmitchell2017leveraging,heetal2020learningtag} and also compute a score $\alpha_i$ for it:
\begin{equation}
\begin{split}
\textbf{c}_i &= \text{FFNN}_\text{s}\big(\textbf{h}_i\big)\\
\alpha_i &= \text{FFNN}_\text{c}\big(\big[\textbf{h}_i, \textbf{c}_i\,\big]\big)
\end{split}
\end{equation}
where $\text{FFNN}_\text{s}$ is another feedforward network mapping from $ \mathbb{R}^d \rightarrow \mathbb{R}^d$. Intuitively, $\textbf{c}_i$ records the information of the local context of $s_i$, and $\alpha_i$ measures the importance of such information. After that, we compute a final knowledge-aware representation $\textbf{f}_i$ for each span $s_i$ as follows:
\begin{equation}
\begin{split}
Z &= \exp{(\alpha_i)} + \sum_{v_z \in C(s_i)} \exp{(\alpha_{iz})} 
\\
\beta_{i} &= \exp{(\alpha_{i})} / Z \;\;\text{and}\;\; \beta_{ij} = \exp{(\alpha_{ij})} / Z \\
\textbf{f}_i &= \beta_{i}\,\textbf{c}_i + \sum_{v_j \in C(s_i)} \beta_{ij} \textbf{n}_{j}
\end{split}
\end{equation}
The attention mechanism is illustrated in Figure \ref{fig:attention_mechanism}.

With the extracted knowledge-aware span representations, we predict the final span graph in a way similar to Eq. \ref{equa:baseline_entity} and Eq. \ref{equa:baseline_relation}:
\begin{equation} \label{equa:final_graph}
\begin{split}
\widehat{\textbf{e}_i} &= \mathtt{Softmax}\big(\text{FFNN}_{\widehat{e}}(\textbf{f}_i)\big) \\
\widehat{\textbf{r}_{ij}} &= \mathtt{Softmax}\big(\text{FFNN}_{\widehat{r}}(\big\lbrack\textbf{f}_i, \textbf{f}_j, \textbf{f}_i \circ \textbf{f}_j\big\rbrack)\big)    
\end{split}
\end{equation}
where $\text{FFNN}_{\widehat{e}}$ is a mapping from $\mathbb{R}^{d} \rightarrow \mathbb{R}^{|E|}$, and $\text{FFNN}_{\widehat{r}}$ is a mapping from $\mathbb{R}^{3 \times d} \rightarrow \mathbb{R}^{|R|}$. $\widehat{\textbf{e}_i}$ is the \textit{final} predicted probability distribution over possible entity types for span $s_i$. $\widehat{\textbf{r}_{ij}}$ is the \textit{final} predicted probability distribution over possible relation types for span pair $\langle s_i, s_j \rangle$.
\subsection{Training} \label{sec:training}

The total loss is computed as:
\begin{equation} \label{eq:loss}
    \mathcal{L}_{total} = (\mathcal{L}^e_1 + \mathcal{L}^r_1) + 2(\mathcal{L}^e_2 + \mathcal{L}^r_2) 
\end{equation}
where $\mathcal{L}^e_\text{*}$ denotes the cross-entropy loss of span classification. $\mathcal{L}^r_\text{*}$ denotes the binary cross-entropy loss of relation classification. $\mathcal{L}^e_1$ and $\mathcal{L}^r_1$ are loss terms for the initial span graph prediction (Eq. \ref{equa:baseline_entity} and Eq. \ref{equa:baseline_relation} of Section \ref{sec:initial_span_graph}). $\mathcal{L}^e_2$ and $\mathcal{L}^r_2$ are loss terms for the final span graph prediction (Eq. \ref{equa:final_graph} of Section \ref{sec:final_graph}). We apply a larger weight score to the loss terms $\mathcal{L}^e_2$ and $\mathcal{L}^r_2$. We train the framework using only ground-truth labels of the entity and relation extraction tasks. We do not make use of any entity linking supervision in this work.
\section{Experiments and Results}
\subsection{Data and Experiments Setup} \label{sec:data_and_exp}
\paragraph{Datasets and evaluation metrics} We evaluate KECI on two benchmark datasets: BioRelEx and ADE. The \textbf{BioRelEx} dataset \cite{khachatrian2019biorelex} consists of 2,010 sentences from biomedical literature that capture binding interactions between proteins and/or biomolecules. BioRelEx has annotations for 33 types of entities and 3 types of relations for binding interactions. The training, development, and test splits contain 1,405, 201, and 404 sentences, respectively. The training and development sets are publicly available. The test set is unreleased and can only be evaluated against using CodaLab \footnote{\,\url{https://competitions.codalab.org/competitions/20468}}. For BioRelEx, we report Micro-F1 scores. The \textbf{ADE} dataset \cite{adedataset} consists of 4,272 sentences extracted from medical reports that describe drug-related adverse effects. Two entity types (\textit{Adverse-Effect} and \textit{Drug}) and a single relation type (\textit{Adverse-Effect}) are pre-defined. Similar to previous work \cite{Eberts2020SpanbasedJE,jietal2020span}, we conduct 10-fold cross-validation and report averaged Macro-F1 scores. All the reported results take overlapping entities into consideration.

\paragraph{Implementation details} We implement KECI using PyTorch \cite{Paszke2019PyTorchAI} and Huggingface's Transformers \cite{wolfetal2020transformers}. KECI uses SciBERT as the Transformer encoder \cite{Beltagy2019SciBERT}. All details about hyperparameters and reproducibility information are in the appendix.

\begin{table}[!t]
\centering
\resizebox{\linewidth}{!}{%
\begin{tabular}{lcc}
\hline
Model & \begin{tabular}[c]{@{}c@{}}Entity\\ (Micro-F1)\end{tabular} & \begin{tabular}[c]{@{}c@{}}Relation\\ (Micro-F1)\end{tabular} \\ \hline
SciIE \shortcite{luanetal2018multi} & 77.90 & 49.60 \\
DYGIEPP + ELMo \shortcite{Bhatt2020BenchmarkingBF} & 81.10 & 55.60 \\
DYGIEPP + BioELMo \shortcite{Bhatt2020BenchmarkingBF} & 82.80 & 54.80 \\\hline
SentContextOnly & 83.98 & 63.90 \\
FlatAttention & 84.32 & 64.23 \\
KnowBertAttention & 85.69 & 65.13\\
Full Model (KECI) & \textbf{87.42} & \textbf{66.09} \\\hline
\end{tabular}%
}
\caption{Overall results (\%) on the development set of BioRelEx.} 
\label{tab:biorelex_dev_results}
\end{table}
\begin{table}[!t]
\centering
\resizebox{0.9\linewidth}{!}{%
\begin{tabular}{lcc}
\hline
Model & \begin{tabular}[c]{@{}c@{}}Entity\\ (Micro-F1)\end{tabular} & \begin{tabular}[c]{@{}c@{}}Relation\\ (Micro-F1)\end{tabular} \\ \hline
SciIE \shortcite{luanetal2018multi}  & 73.56 & 50.15 \\
Second Best Model & 82.76 & 62.18 \\ \hline
Full Model (KECI) & \textbf{87.35} & \textbf{67.09} \\\hline
\end{tabular}%
}
\caption{Overall results (\%) on the test set of BioRelEx (from the leaderboard as of January 20th, 2021).}
\label{tab:biorelex_test_results}
\end{table}

\paragraph{Baselines for comparison} In addition to comparing our method with state-of-the-art methods on the above two datasets, we implement the following baselines for further comparison and analysis:
\begin{enumerate}[topsep=0pt,itemsep=-1ex,partopsep=1ex,parsep=1ex]
    \item \textbf{SentContextOnly}: This baseline does not use any \textit{external knowledge}. It uses only the local sentence context for prediction. It extracts the final output directly from the predictions obtained using Eq. \ref{equa:baseline_entity} and Eq. \ref{equa:baseline_relation}.
    \item \textbf{FlatAttention}: This baseline does not rely on \textit{collective inference}. It does not integrate any global relational information into mention and entity representations. Each $\textbf{h}_i$ mentioned in Sec. \ref{sec:initial_span_graph} is set to be $\textbf{s}_i$ (Eq. \ref{eq:base_span_emb}), and each $\textbf{v}_i$ mentioned in Sec. \ref{sec:external_knowledge} is set to be $\textbf{v}_i^0$. Then, the prediction of the final span graph is the same as described in Sec. \ref{sec:final_graph}.
    \item \textbf{KnowBertAttention}: This baseline uses the Knowledge Attention and Recontextualization (KAR) mechanism of KnowBert \cite{knowbert}, \textit{a state-of-the-art knowledge-enhanced language model}. The baseline first uses SciBERT to construct initial token-level representations. It then uses the KAR mechanism to inject external knowledge from UMLS into the token-level vectors. Finally, it embeds text spans into feature vectors (Eq. \ref{eq:base_span_emb}) and uses the span representations to extract entities and relations in one pass (similar to Eq. \ref{equa:final_graph}).
\end{enumerate}
For fair comparison, all the baselines use SciBERT as the Transformer encoder.


A major difference between KECI and KnowBertAttention \cite{knowbert} is that KECI explicitly builds and extracts information from a multi-relational graph structure of the candidate entity mentions before the knowledge fusion process. In contrast, KnowBertAttention only uses SciBERT to extract features from the candidate entity mentions. Therefore, KnowBertAttention only takes advantage of the entity-entity co-occurrence information. On the other hand, KECI integrates more fine-grained global relational information (e.g., the binding interactions shown in Figure \ref{fig:approach_overview}) into the mention representations. This difference makes KECI achieve better overall performance, as to be discussed next.


\subsection{Overall Results}

\begin{table}[!t]
\centering
\resizebox{\linewidth}{!}{%
\begin{tabular}{lcc}
\hline
Model & \begin{tabular}[c]{@{}c@{}}Entity\\ (Macro-F1)\end{tabular} & \begin{tabular}[c]{@{}c@{}}Relation\\ (Macro-F1)\end{tabular} \\ \hline
Relation-Metric \shortcite{Tran2019NeuralML} & 87.11 & 77.29\\
SpERT \shortcite{Eberts2020SpanbasedJE} & 89.28 & 78.84\\
$\text{SPAN}_{\text{Multi-Head}}$ \shortcite{jietal2020span} & 90.59 & 80.73\\\hline
SentContextOnly & 88.13 & 77.23 \\
FlatAttention & 89.16 & 78.81\\
KnowBertAttention & 90.08 & 79.95 \\
Full Model (KECI) & \textbf{90.67} & \textbf{81.74} \\\hline
\end{tabular}%
}
\caption{Overall results (\%) on the ADE dataset.}
\label{tab:ade_overall_results}
\end{table}
\begin{table}[!t]
\centering
\resizebox{\linewidth}{!}{%
\begin{tabular}{lcc}
\hline
Ablation setting & \begin{tabular}[c]{@{}c@{}}Entity\\ (Micro-F1)\end{tabular} & \begin{tabular}[c]{@{}c@{}}Relation\\ (Micro-F1)\end{tabular} \\ \hline
Full Model (KECI) & \textbf{87.42} & \textbf{66.09} \\\hline
$\bullet$ w/o external knowledge & 83.98* & 63.90* \\
$\bullet$ w/o collective inference & 84.32* & 64.23* \\
$\bullet$ w/o the bidirectional GCN & 84.76* & 64.25*  \\
$\bullet$ w/o the relational GCN & 85.14* & 65.32*\\
$\bullet$ w/o the pretrained UMLS vectors & 86.25${}^\dagger$ & 65.29* \\
$\bullet$ w/o the UMLS definition vectors & 86.76${}^\dagger$ & 65.45${}^\dagger$ \\\hline
\end{tabular}%
}
\caption{Results (\%) of ablation experiments on the development set of BioRelEx. We use the symbols * and $\dagger$ to indicate statistical significance with 95\% and 90\% confidence levels respectively (compared to KECI).} 
\label{tab:biorelex_ablation_study}
\end{table}

Table \ref{tab:biorelex_dev_results} and Table \ref{tab:biorelex_test_results} show the overall results on the development and test sets of BioRelEx, respectively. Compared to SentContextOnly, KECI achieves much higher performance. This demonstrates the importance of incorporating external knowledge for biomedical information extraction. KECI also outperforms the baseline FlatAttention by a large margin, which shows the benefit of collective inference. In addition, we see that our model performs better than the baseline KnowBertAttention. Finally, at the time of writing, KECI achieves the first position on the BioRelEx leaderboard \footnote{\,\url{https://competitions.codalab.org/competitions/20468}}.

\renewcommand{\arraystretch}{1.75}
\begin{table*}[!ht]
\centering
\footnotesize
\begin{tabular}{|m{1.25cm}|m{6.5cm}|m{6.5cm}|}
\hline
Datasets & Top 3 types with the \textcolor{red}{lowest} avg. attention scores & Top 3 types with the \textcolor{blue}{highest} avg. attention scores \\ \hline
BioRelEx & Diagnostic Procedure (0.04); Activity (0.05); Plant (0.05) &  Amino Acid, Peptide, or Protein (0.32); Enzyme (0.32); Molecular Function (0.36) \\ \hline
ADE & Intellectual Product (0.15); Idea or Concept (0.19); Temporal Concept (0.19) & Antibiotic (0.78); Organic Chemical (0.79); Nucleic Acid, Nucleoside, or Nucleotide (0.87) \\ \hline
\end{tabular}
\caption{Average attention scores of different UMLS semantic types.}
\label{tab:attention_pattern_table}
\end{table*}
\renewcommand{\arraystretch}{1.25}
\begin{table*}[!ht]
\centering
\small
\begin{tabular}{|m{6.25cm}|c|c|}
\hline
\textbf{Input Sentence} & \textbf{Initial Span Graph} & \textbf{Final Span Graph}\\\hline
\#1: \textit{Despite the low dosage of \textcolor{drugspancolor}{warfarin}, international normalized ratio (INR) was markedly elevated from 1.15 to 11.28 for only 4 days, and \textcolor{effectspancolor}{bleeding symptoms} concurrently developed.}
& \includegraphics[scale=0.28, valign=m, margin=0.5em]{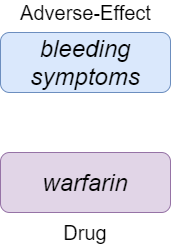} 
& \includegraphics[scale=0.28, valign=m, margin=0.5em]{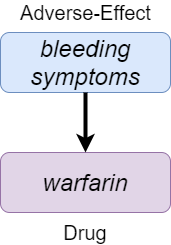}\\ 
\hline
\#2: \textit{A 25-year-old woman sought medical attention because of iliocaval manifestations of \textcolor{effectspancolor}{retroperitoneal fibrosis} while she was taking \textcolor{drugspancolor}{methysergide}}. & \includegraphics[scale=0.28, valign=m, margin=0.5em]{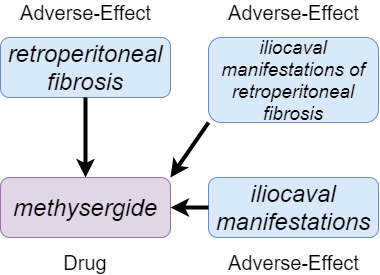} &
\includegraphics[scale=0.28, valign=m, margin=0.5em]{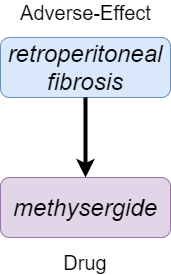} \\\hline
\#3: TITLE: \textit{\textcolor{effectspancolor}{Acute abdomen} due to \textcolor{effectspancolor}{endometriosis} in a premenopausal woman taking \textcolor{drugspancolor}{tamoxifen}.} & \includegraphics[scale=0.28, valign=m, margin=0.5em]{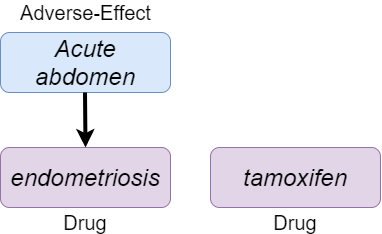} & \includegraphics[scale=0.28, valign=m, margin=0.5em]{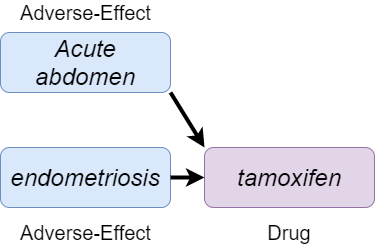} \\\hline
\end{tabular}
\caption{Examples showing how external knowledge improves the quality of extracted span graphs. Edges represent relations of type \textit{Adverse-Effect}. Only relations with predicted probabilities of at least 0.5 are shown.}
\label{tab:ade_qualitative_examples}
\end{table*}


Table \ref{tab:ade_overall_results} shows the overall results on ADE. KECI again outperforms all the baselines and state-of-the-art models such as SpERT \cite{Eberts2020SpanbasedJE} and $\text{SPAN}_{\text{Multi-Head}}$ \cite{jietal2020span}. This further confirms the effectiveness of our framework. 

Overall, the two datasets used in this work focus on two very different subareas of the biomedical domain, and KECI was able to push the state-of-the-art results of both datasets. This indicates that our proposed approach is highly generalizable.
\subsection{Ablation Study}

Table \ref{tab:biorelex_ablation_study} shows the results of ablation studies we did on the development set of the BioRelEx benchmark. We compare our full model against several partial variants. The variant [w/o external knowledge] is the same as the baseline SentContextOnly, and the variant [w/o collective inference] is the same as the baseline FlatAttention (Section \ref{sec:data_and_exp}). For the variant [w/o the bidirectional GCN], we simply set each $\textbf{h}_i$ mentioned in Section \ref{sec:initial_span_graph} to be $\textbf{s}_i$. Similarly, for the variant [w/o the relational GCN], we set each $\textbf{v}_i$ in Section \ref{sec:external_knowledge} to be $\textbf{v}_i^0$. The last two variants are related to the initialization of each vector $\textbf{v}_i^0$.

We see that all the partial variants perform worse than our full model. This shows that each component of KECI plays an important role.

\subsection{Attention Pattern Analysis} \label{sec:attn_pattern}

There is no gold-standard set of correspondences between the entity mentions in the datasets and the UMLS entities. Therefore, we cannot directly evaluate the entity linking performance of KECI. However, for each UMLS semantic type, we compute the average attention weight that an entity of that type gets assigned (Table \ref{tab:attention_pattern_table}). Overall, we see that KECI typically pays the most attention to the relevant informative entities while ignoring the irrelevant ones. 

\subsection{Qualitative Analysis}

Table \ref{tab:ade_qualitative_examples} shows some examples from the ADE dataset that illustrate how incorporating external knowledge can improve the performance of joint biomedical entity and relation extraction.

In the first example, initially, there is no edge between the node ``bleeding symptoms'' and the node ``warfarin'', probably because of the distance between their corresponding spans in the original input sentence. However, KECI can link the term ``warfarin'' to a UMLS entity (CUI: C0043031), and the definition in UMLS says that warfarin is a type of anticoagulant that prevents the formation of blood clots. As the initial feature vector of each entity contains the representation of its definition (Sec. \ref{sec:external_knowledge}), KECI can recover the missing edge.

In the second example, the initial span graph is predicted to have three entities of type \textit{Adverse-Effect}, which correspond to three different overlapping text spans. Among these three, only ``retroperitoneal fibrosis'' can be linked to a UMLS entity. It is also evident from the input sentence that one of these spans is related to ``methysergide''. As a result, KECI successfully removes the other two unlinked span nodes to create the final span graph.

In the third example, probably because of the phrase ``due to'', the node ``endometriosis'' is initially predicted to be of type \textit{Drug}, and the node ``acute abdomen'' is predicted to be its \textit{Adverse-Effect}. However, KECI can link the term ``endometriosis'' to a UMLS entity of semantic type \textit{Disease or Syndrome}. As a result, the system can correct the term's type and also predict the right edges for the final span graph.

Finally, we also examined the errors made by KECI. One major issue is that MetaMap sometimes fails to return any candidate entity from UMLS for an entity mention. We leave the extension of this work to using multiple KBs as future work.



\section{Related Work}


Traditional pipelined methods typically treat entity extraction and relation extraction as two separate tasks \cite{zelenkoetal2002kernel,zhouetal2005exploring,chanroth2011exploiting}. Such approaches ignore the close interaction between named entities and their relation information and typically suffer from the error propagation problem. To overcome these limitations, many studies have proposed joint models that perform entity extraction and relation extraction simultaneously \cite{roth2007global,liji2014incremental,li2017neural,zhengetal2017joint,Bekoulis2018JointER,bekoulisetal2018adversarial,waddenetal2019entity,fuetal2019graphrel,luanetal2019general,ijcai2020558,wanglu2020two,Li2020BiosemanticRE,linetal2020joint}. Particularly, span-based joint extraction methods have gained much popularity lately because of their ability to detect overlapping entities. For example, \newcite{Eberts2020SpanbasedJE} propose SpERT, a simple but effective span-based model that utilizes BERT as its core. The recent work of \newcite{jietal2020span} also closely follows the overall architecture of SpERT but differs in span-specific and contextual semantic representations. Despite their impressive performance, these methods are not designed specifically for the biomedical domain, and they do not utilize any external knowledge base. To the best of our knowledge, our work is the first span-based framework that utilizes external knowledge for joint entity and relation extraction from biomedical text. 

Biomedical event extraction is a closely related task that has also received a lot of attention from the research community \cite{poonvanderwende2010joint,kimetal2013genia,vsspatchigollaetal2017biomedical,raoetal2017biomedical,espinosaetal2019search,lietal2019biomedical,wangetal2020biomedical,huangetal2020biomedical,ramponietal2020biomedical,10.1145/3372328}. Several studies have proposed to incorporate external knowledge from domain-specific KBs into neural models for biomedical event extraction. For example, \newcite{lietal2019biomedical} incorporate entity information from Gene Ontology into tree-LSTM models. However, their approach does not explicitly use any external relational information. Recently, \newcite{huangetal2020biomedical} introduce a framework that uses a novel Graph Edge conditioned Attention Network (GEANet) to utilize domain knowledge from UMLS. In the framework, a global KG for the entire corpus is first constructed, and then a sentence-level KG is created for each individual sentence in the corpus. Our method of KG construction is more flexible as we directly create a KG for each input text. Furthermore, the work of \newcite{huangetal2020biomedical} only deals with event extraction and assumes that gold-standard entity mentions are provided at inference time.

Some previous work has focused on integrating external knowledge into neural architectures for other tasks, such as reading comprehension \cite{mihaylovfrank2018knowledgeable}, question answering~\cite{Pan2019}, natural language inference \cite{sharmaetal2019incorporating}, and conversational modeling \cite{parthasarathipineau2018extending}. Different from these studies, our work explicitly emphasizes the benefit of collective inference using global relational information. 

Many previous studies have also used GNNs for various IE tasks \cite{Nguyen2018GraphCN,liuetal2018jointly,SubburathinamEMNLP2019,zengetal2021gene,zhangji2021abstract}. Many of these methods use a dependency parser or a semantic parser to construct a graph capturing global interactions between tokens/spans. However, parsers for specialized biomedical domains are expensive to build. KECI does not rely on such expensive resources.




\section{Conclusions and Future Work}

In this work, we propose a novel span-based framework named KECI that utilizes external domain knowledge for joint entity and relation extraction from biomedical text. Experimental results show that KECI is highly effective, achieving new state-of-the-art results on two datasets: BioRelEx and ADE. Theoretically, KECI can take an entire document as input; however, the tested datasets are only sentence-level datasets. In the future, we plan to evaluate our framework on more document-level datasets. We also plan to explore a broader range of properties and information that can be extracted from external KBs to facilitate biomedical IE tasks. Finally, we also plan to apply KECI to other information extraction tasks \cite{ligaiasmkbp2020,laietal2021context,wenetal2021resin}.




\section*{Acknowledgement}
We thank the three reviewers and the Area Chair for their insightful comments and suggestions. 
This research is based upon work supported by the Molecule Maker Lab Institute: An AI Research Institutes program supported by NSF under Award No. 2019897, NSF No. 2034562, U.S. DARPA KAIROS Program No. FA8750-19-2-1004, the Office of the Director of National Intelligence (ODNI), Intelligence Advanced Research Projects Activity (IARPA), via contract No. FA8650-17-C-9116. Any opinions, findings and conclusions or recommendations expressed in this document are those of the authors and should not be interpreted as representing the official policies, either expressed or implied, of the U.S. Government. The U.S. Government is authorized to reproduce and distribute reprints for Government purposes notwithstanding any copyright notation here on.

\bibliographystyle{acl_natbib}
\bibliography{anthology,acl2021}

\appendix
\section{Reproducibility Checklist} \label{sec:reproducibility_checklist}
In this section, we present the reproducibility information of the paper. We are planning to make the code publicly available after the paper is reviewed.

\paragraph{Implementation Dependencies Libraries} Pytorch 1.6.0 \cite{Paszke2019PyTorchAI}, Transformers 4.0.0 \cite{wolfetal2020transformers}, DGL 0.5.3\footnote{\url{https://www.dgl.ai/}}, Numpy 1.19.1 \cite{Harris2020ArrayPW}, CUDA 10.2.

\paragraph{Computing Infrastructure} The experiments were conducted on a server with Intel(R) Xeon(R) Gold 5120 CPU @ 2.20GHz and NVIDIA Tesla V100 GPUs. The allocated RAM is 187G. GPU memory is 16G.

\paragraph{Datasets} The BioRelEx dataset \cite{khachatrian2019biorelex} is available at \url{https://github.com/YerevaNN/BioRelEx}. The ADE dataset \cite{adedataset} can be downloaded by using the script at \url{https://github.com/markus-eberts/spert}.

\paragraph{Average Runtime} Table \ref{tab:average_runtime} shows the estimated average run time of our full model.

\paragraph{Number of Model Parameters} The number of parameters in a full model trained on BioRelEx is about 121.0M parameters. The number of parameters in a full model trained on ADE is about 119.9M parameters.

\paragraph{Hyperparameters of Best-Performing Models} The span length limit $L$ is set to be 20 tokens. Note that the choice of $L$ only has some noticeable effects on the training time of KECI during the first epoch. KECI with randomly initialized parameters may include many non-relevant spans in the initial span graph. However, after a few training iterations, KECI typically can filter out most non-relevant spans. The pruning parameter $\lambda$ is set to be 0.5. All of our models use SciBERT as the Transformer encoder \cite{Beltagy2019SciBERT}. We use two different learning rates, one for the lower pretrained Transformer encoder and one for the upper layers. Table \ref{tab:best_params} summarizes the hyperparameter configurations of best-performing models.

\paragraph{Expected Validation Performance} The main paper has the results on the dev set of BioRelEx. For ADE, as in previous work, we conduct a 10-fold cross validation.

\paragraph{Hyperparameter Tuning Process}  We experimented with the following range of possible values: \{16, 32\} for batch size, \{2e-5, 3e-5, 4e-5, 5e-5\} for lower learning rate, \{1e-4, 2e-4, 5e-4\} for upper learning rate, and \{50, 100\} for number of training epochs. For each particular set of hyperparameters, we repeat training for 3 times and compute the average performance. 

\begin{table}[!h]
\centering
\resizebox{0.75\linewidth}{!}{%
\begin{tabular}{c|c|c}
Dataset & \begin{tabular}[c]{@{}c@{}}One Training \\ Epoch\end{tabular} & \begin{tabular}[c]{@{}c@{}}Evaluation \\ (Dev Set)\end{tabular} \\\hline
BioRelEx &  337.51 seconds & 35.38 seconds \\\hline
ADE & 712.89 seconds & 52.39 seconds
\end{tabular}%
}
\caption{Estimated average runtime of our full model.}
\label{tab:average_runtime}
\end{table}

\begin{table}[!h]
\centering
\resizebox{0.8\linewidth}{!}{%
\begin{tabular}{c|c|c}
Hyperparameters & BioRelEx & ADE \\\hline
Lower Learning Rate & 5e-05 & 5e-05 \\\hline
Upper Learning Rate & 2e-04 & 1e-04 \\\hline
Batch Size & 32 & 32 \\\hline
Number Epochs & 50 & 50 \\\hline
\end{tabular}%
}
\caption{Hyperparameters for best-performing models.}
\label{tab:best_params}
\end{table}

\end{document}